\documentclass[runningheads]{llncs}

\usepackage[final,year=2026]{eccv}

\usepackage{eccvabbrv}

\usepackage{graphicx}
\usepackage{booktabs}
\usepackage{multirow}
\usepackage{iftex}
\ifPDFTeX
  \usepackage[accsupp]{axessibility}
\fi

\usepackage{hyperref}

\usepackage{orcidlink}

\hypersetup{
  hidelinks,
  pdftitle={Seeing as Humans Do: Learning from Motion to Segment Anything Without Supervision},
  pdfauthor={Weijian Jian; Xiaoyue Zhang; Bin Xiao; Chunyu Xie; Yixiao He; Yutao Liu; Dawei Leng; Yuhui Yin},
  pdfsubject={Computer Vision -- ECCV 2026},
  pdfkeywords={Unsupervised Learning, Segment Anything, MoSA}
}

\begin{document}

\title{Seeing as Humans Do: Learning from Motion to Segment Anything Without Supervision}

\titlerunning{Seeing as Humans Do}

\author{
Weijian Jian\inst{1,\ast}\orcidlink{0000-0002-8055-8658} \and
Xiaoyue Zhang\inst{2,\ast}\orcidlink{0009-0002-2432-079X} \and
Bin Xiao\inst{3}\orcidlink{0000-0003-1992-4214} \and
Chunyu Xie\inst{1}\orcidlink{0009-0002-6607-8209} \and
Yixiao He\inst{4}\orcidlink{0009-0006-2746-1327} \and
Yutao Liu\inst{1}\orcidlink{0009-0001-8049-0163} \and
Dawei Leng\inst{1,\boxtimes}\orcidlink{0009-0000-5461-1681} \and
Yuhui Yin\inst{1}\orcidlink{0009-0000-8761-7334}
}

\authorrunning{W.~Jian et al.}

\institute{
360 AI Research
\and
Independent Researcher
\and
University of Ottawa
\and
Beijing University of Posts and Telecommunications}

\maketitle

\begingroup
\renewcommand{\thefootnote}{}
\footnotetext{$^\ast$Equal contribution, $^\boxtimes$Corresponding author.\\
Author manuscript of the paper published in \emph{Computer Vision -- ECCV 2026},
LNCS 17014, pp.~600--616. The version of record is available at
\url{https://doi.org/10.1007/978-3-032-37232-1_33}.}
\endgroup

\begin{abstract}

The Segment Anything Model (SAM) relies heavily on massive manual annotations, creating a fundamental bottleneck for model scaling. While unsupervised methods attempt to learn object concepts from motion, they typically overfit to moving entities, lacking both multi-granularity understanding and the ability to generalize to static objects. To overcome this, we introduce \textbf{Mo}tion-Grounded \textbf{S}egment \textbf{A}nything (\textbf{MoSA}), a highly scalable unsupervised framework that learns a transferable objectness prior from unlabeled videos. MoSA operates in three progressive stages: (1) automatically generating multi-granularity motion pseudo-labels from large-scale video data; (2) training a \textbf{P}erceptual \textbf{G}rouping \textbf{M}odel (\textbf{PGM}) via contrastive learning to internalize a generalized, appearance-driven concept of objects; and (3) transferring this learned prior into a prompt-guided architecture for segment-anything-style inference on images. Extensive zero-shot evaluations across seven challenging benchmarks (e.g., COCO and ADE20K) demonstrate that MoSA significantly outperforms existing unsupervised methods. Notably, despite using zero manual annotations, MoSA achieves segmentation performance comparable to the fully supervised SAM. Our findings reveal that harnessing large-scale unlabeled motion is a feasible and highly scalable alternative to annotation-driven segment-anything pipelines.

Code is available at \url{https://github.com/360CVGroup/MoSA}.

\end{abstract}

\section{Introduction}
\label{sec:intro}
\begin{figure*}[h]
     \centering
     \includegraphics[width=1.0\textwidth]{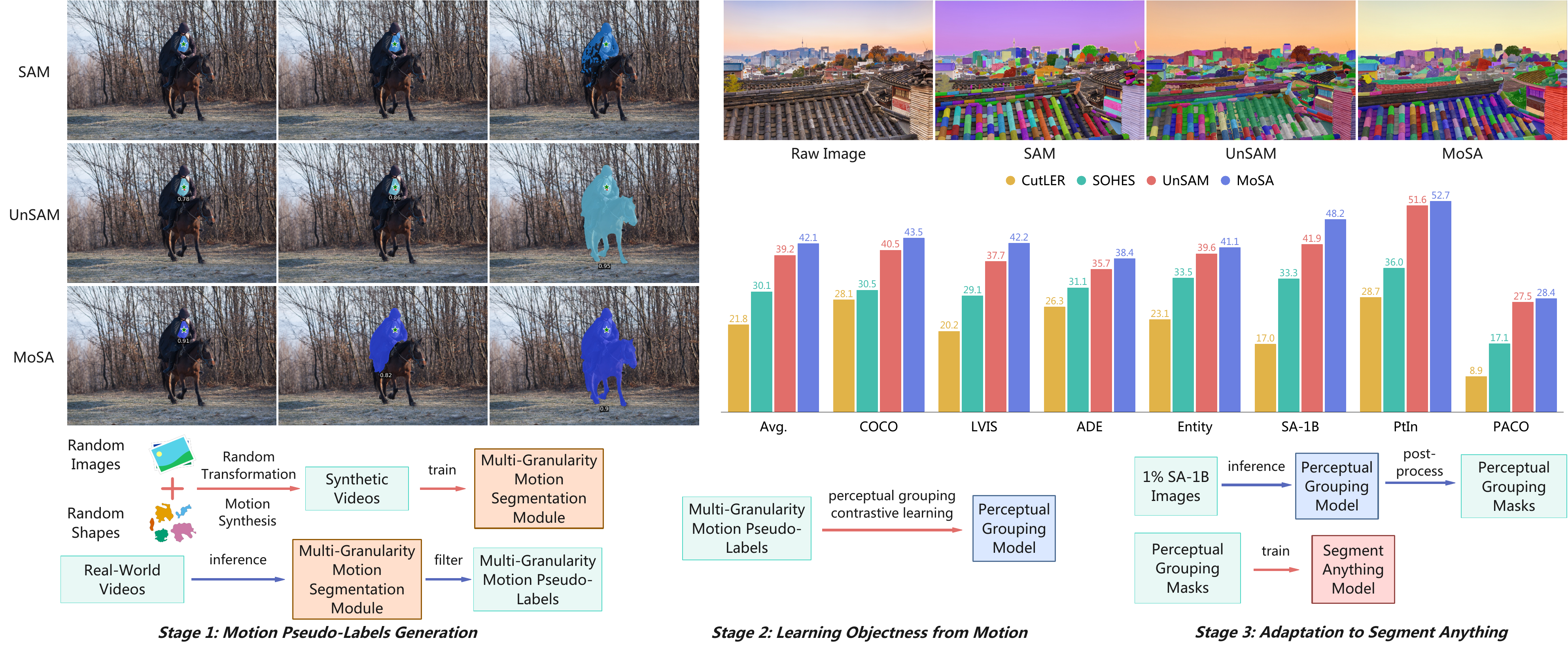}
     \caption{Overview of our proposed \textbf{MoSA}. The \textbf{top-left} and \textbf{top-right} panels respectively show qualitative comparisons of MoSA against the fully supervised SAM~\cite{SAM} and the unsupervised UnSAM~\cite{UnSAM} on point-based promptable segmentation and whole-image segmentation. The middle-right panel shows that MoSA achieves SOTA performance compared to other unsupervised segmentation methods. The \textbf{bottom} panel illustrates the pipeline of MoSA's three-stage unsupervised framework. Note that UnSAM is not strictly unsupervised, as it utilizes the supervised CascadePSP~\cite{CascadePSP} refiner.}
     \label{fig:title_image}
 \end{figure*}

The Segment Anything Model (SAM)~\cite{SAM} has established a new standard for zero-shot generalization in computer vision. However, its success heavily relies on the massive, manually annotated SA-1B dataset. This dependence on costly supervision creates a fundamental bottleneck for further model scaling. Consequently, annotation-free unsupervised learning has emerged as a crucial direction for scalable, general-purpose segmentation.

Existing unsupervised segmentation methods generally fall into two categories: image-based and video-based approaches. Image-based methods, such as CutLER~\cite{CutLER} and UnSAM~\cite{UnSAM}, leverage self-supervised representations (e.g., DINO~\cite{DINO}) to perform feature clustering or pseudo-label generation, achieving competitive unsupervised segmentation performance. Nevertheless, these approaches infer objectness primarily from static appearance cues, such as texture, color, and shape, overlooking the cognitive priors humans develop through continuous observation of the dynamic world.

Research in developmental psychology, particularly Gestalt theory~\cite{kohler1967gestalt, koffka2013principles}, suggests that the human ability to group objects is largely driven by motion cues. Video-based methods naturally exploit these motion cues. Early approaches typically focus on detecting moving foreground regions, often producing binary masks without explicitly modeling multiple object instances or detailed structures~\cite{yang2019unsupervised,yang2021self,lee2023unsupervised}. Recent advances extend unsupervised video object segmentation (U-VOS) toward instance- and scene-level understanding. For example, OCLR~\cite{OCLR} proposes an object-centric model trained on synthetic data to discover and track multiple moving objects. Other methods rely on videos during training but can perform inference on static images without requiring motion information. Methods such as RCF~\cite{lian2023bootstrapping} and DyStaB~\cite{yang2021dystab} exploit motion-induced grouping to generate object masks, often combining motion and appearance refinement within bootstrapping frameworks. Furthermore, approaches like DIOD~\cite{kara2024diod} integrate motion-guided discovery with self-distillation to improve pseudo-label quality. Although these methods prove that motion is a powerful supervisory signal for object concepts, they face two major limitations. First, they easily overfit to moving objects, struggling to generalize to static or weakly moving entities (e.g., the roof tiles in Fig.~\ref{fig:title_image}). Second, they are largely confined to instance-level segmentation, lacking multi-granularity understanding. To truly emulate human perception, a scalable model must generalize from moving entities to \textit{anything} in static scenes, expanding its object vocabulary from growing unlabeled videos.

Motivated by this insight, we introduce \textbf{Mo}tion-Grounded \textbf{S}egment \textbf{A}nything (\textbf{MoSA}), a three-stage unsupervised framework designed to learn a transferable, general-purpose segmentation capability from unlabeled videos (see Fig.~\ref{fig:title_image}). In the first stage, we develop a \textbf{M}ulti-\textbf{G}ranularity \textbf{M}otion \textbf{S}egmentation (\textbf{MGMS}) pipeline. Pre-trained on synthetic data and subsequently applied to real-world unlabeled videos, MGMS generates high-quality motion-based pseudo-labels at scale without manual annotation. In the second stage, we train a \textbf{P}erceptual \textbf{G}rouping \textbf{M}odel (\textbf{PGM}) using these motion pseudo-labels. This stage encourages the model to internalize a generalized, appearance-driven notion of objectness that no longer depends on motion cues. Finally, an efficient adaptation stage transfers the learned objectness prior to a prompt-guided, high-resolution segmentation architecture, enabling strong zero-shot segment-anything performance. We conduct extensive evaluations on seven challenging segmentation benchmarks, including COCO~\cite{lin2014microsoft} and ADE20K~\cite{zhou2019semantic}. Experimental results demonstrate that MoSA consistently outperforms existing unsupervised segmentation methods across all benchmarks. Notably, despite being trained entirely without manual annotations, MoSA achieves zero-shot segmentation performance comparable to SAM, which relies on large-scale supervised data. These findings suggest that learning general object perception from large-scale unlabeled motion is both feasible and scalable, offering a promising path toward foundational general-purpose segmentation. Our main contributions are as follows:
\begin{itemize}
    \item We propose \textbf{MoSA}, a motion-grounded segmentation framework that learns from unlabeled videos to acquire a general-purpose, multi-granularity segmentation capability without manual annotation.
    \item We introduce a \textbf{P}erceptual \textbf{G}rouping \textbf{M}odel (\textbf{PGM}) with a contrastive learning objective that allows the model to generalize beyond motion-specific cues, facilitating the discovery and segmentation of unseen static objects.
    \item MoSA achieves new state-of-the-art performance in unsupervised segmentation across extensive experiments, proving that learning general objectness from motion is a scalable and practical route to developing foundational vision models.
\end{itemize}

\section{Related Work}
\label{sec:related}

\subsubsection{Unsupervised Image Segmentation}
Unsupervised image segmentation aims to automatically identify and segment objects from static images without manual annotations~\cite{hu2018unsupervised,wang2022freesolo}. Methods such as LOST~\cite{simeoni2021localizing} and TokenCut~\cite{wang2023tokencut} leverage patch-level features from pretrained vision transformers (e.g., DINO family~\cite{DINO,DINOv2,DINOv3}) to localize salient objects via feature affinity and graph-based partitioning. CutLER~\cite{CutLER} introduces a cut-and-learn pipeline that discovers pseudo-masks and trains segmentation models in a self-supervised manner, demonstrating strong multi-instance detection and segmentation capabilities. UnSAM~\cite{UnSAM} adopts a divide-and-conquer strategy that combines hierarchical grouping and mask refinement to generate multi-granularity pseudo-labels, enabling both automatic and promptable segmentation. A common thread in these image-based approaches is their reliance on DINO features and the subsequent use of hierarchical clustering to generate pseudo-labels. However, they may lack the robust, world-grounded prior knowledge that humans develop through extensive, long-term observation.

\subsubsection{Motion-to-Objectness Bootstrapping}
A growing body of research leverages motion as a natural supervisory signal for self-supervised object discovery. The standard paradigm extracts motion segments from videos to train static image models, allowing the concept of ``objects'' to generalize beyond initially moving regions. For instance, DyStaB~\cite{yang2021dystab} partitions the motion field by minimizing the mutual information between segments, then employs a dynamic-static bootstrapping strategy that iteratively alternates between motion-based segmentation and static model learning. RCF~\cite{lian2023bootstrapping} combines relaxed common-fate grouping with appearance refinement to generate object masks. Furthermore, Bao \etal~\cite{bao2022discovering} utilize a slot-attention framework to discover independently moving objects, while DIOD~\cite{kara2024diod} integrates motion guidance with self-distillation to enhance pseudo-label quality. However, existing methods remain confined to narrow, specialized scenarios. Their training objectives focus heavily on extracting precise instance masks rather than learning a universal, transferable ``object prior'' necessary for foundational ``segment-anything'' capabilities. Furthermore, they are strictly restricted to instance-level predictions and cannot comprehend objects across multiple granularities (e.g., distinguishing a part from a whole). Consequently, their learned representations suffer from a severe \textit{motion bias}---they overfit to motion-salient regions and fail on static or rarely moving objects. Ultimately, building a scalable, general-purpose segmentation model solely from motion cues remains an open challenge.

\begin{figure}[t]
    \centering
    \includegraphics[width=0.85\textwidth]{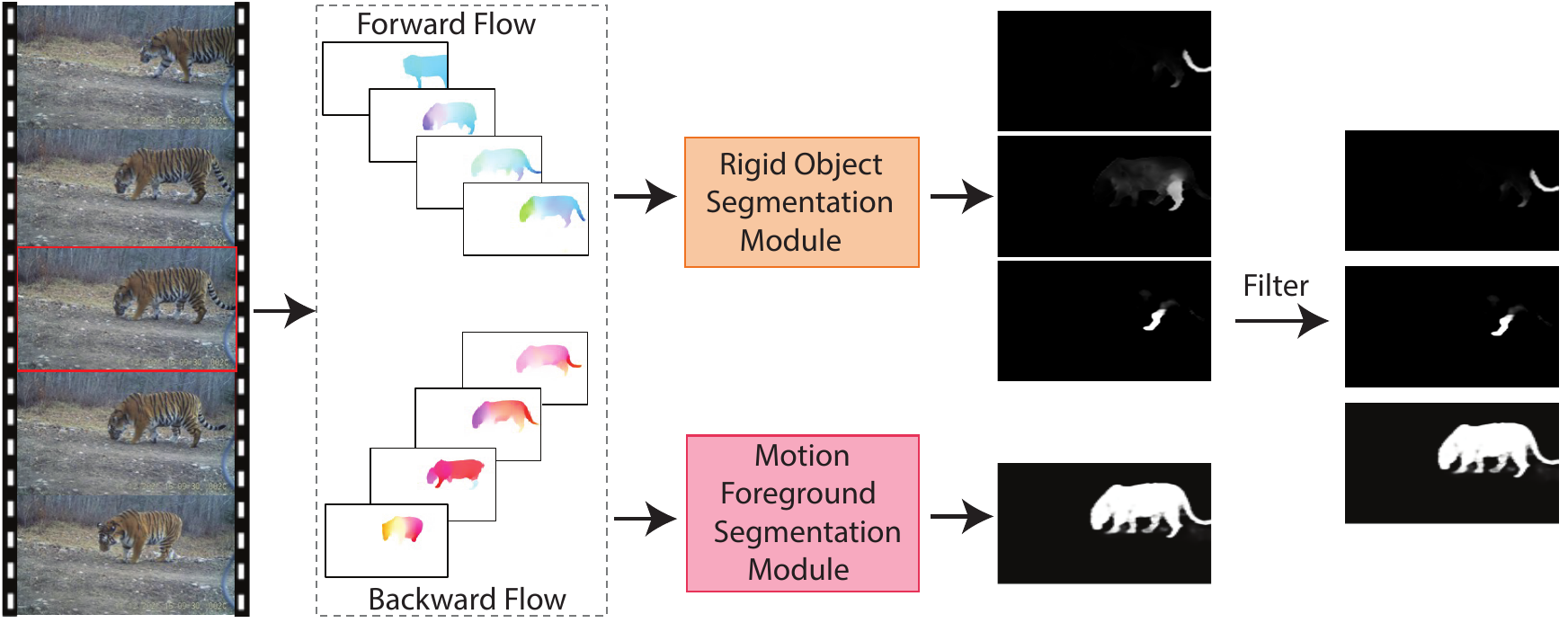}
    \caption{An overview of our Multi-Granularity Motion Segmentation (MGMS) pipeline for generating high-quality pseudo-labels from unlabeled videos. The pipeline first extracts bidirectional optical flow from videos and processes it through two parallel modules: the Rigid Object Segmentation Module and the Motion Foreground Segmentation Module. These generate masks for rigid parts and the moving foreground of the central frame (highlighted in red). Finally, masks are filtered and processed with NMS to yield the final multi-granularity pseudo-labels.}
    \label{fig:stage_1}
\end{figure}

\section{Motion Pseudo-Label Generation}
\label{sec:Pseudo_Labels_Generation}

To learn object concepts from motion without manual supervision, MoSA begins with a \textbf{M}ulti-\textbf{G}ranularity \textbf{M}otion \textbf{S}egmentation (\textbf{MGMS}) pipeline that generates high-quality pseudo-labels from unlabeled videos. Within the MGMS pipeline, motion segmentation modules are first trained on synthetic video data. These modules are subsequently applied to large-scale real-world videos to produce a large set of candidate masks (see Fig.~\ref{fig:stage_1}). Finally, a dedicated quality assessment metric is introduced to filter the generated candidates, resulting in a high-quality, multi-granularity motion pseudo-label dataset for subsequent-stage training.

\subsection{Synthetic Video Generation}
\label{ssec:synthetic}
Segmenting moving objects from real-world videos is inherently challenging due to confounding factors such as complex camera motion, occlusions, and diverse object dynamics (e.g., rigid versus non-rigid motion). To mitigate these challenges and provide clean supervisory signals, we introduce a video synthesis pipeline to train our motion segmentation modules. Inspired by~\cite{lamdouar2020}, we generate synthetic videos with accurately constructed ground-truth masks. We extend the original framework to support multiple objects and complex occlusions. Specifically, for each synthetic video, a background image is sampled as the background texture, and random affine transformations are applied to simulate camera motion. We then generate multiple foreground objects, represented as randomly shaped and textured patches. Each object moves independently along a smooth B-spline trajectory and undergoes random perspective transformations, with inter-frame smoothness constraints to mimic natural motion and deformation. Although synthetic videos lack semantic realism, they provide unambiguous motion cues and pixel-accurate masks, making them well-suited for training general-purpose motion segmentation models.

\subsection{Multi-Granularity Motion Segmentation}
\label{ssec:modules}
To capture motion at varying levels of detail, we design a pipeline with two complementary modules, both trained on our synthetic data. The first module focuses on instance-level segmentation of rigidly moving objects, while the second identifies the entire moving foreground. This multi-granularity strategy ensures a comprehensive capture of diverse motion patterns. Further training details for each module are provided in Appendix~\ref{app:mgms_training}.

\subsubsection{Rigid Object Segmentation Module (ROSM)}
This module is designed to segment individual objects that undergo coherent motion. In our context, ``rigid motion'' refers to any motion pattern that can be well-approximated by a single perspective transformation. This assumption holds not only for truly rigid objects (e.g., a car), but also for instance parts (e.g., a person's arm) or even static elements exhibiting parallax due to camera motion. Our synthetic data, where each object follows a distinct transformation, provides an ideal training ground for this task. We formulate this as an instance segmentation problem, adapting the SOLOv2 framework~\cite{wang2020solov2} to operate on bidirectional optical flow. For a given flow field, the module outputs a set of instance masks, each corresponding to a distinct moving entity, along with a confidence score.

\subsubsection{Motion Foreground Segmentation Module (MFSM)}
While the rigid object module excels at segmenting coherent motion, it struggles with complex, non-rigid deformations, such as a walking person whose limbs move semi-independently. To address this, we introduce a module that performs a simpler, yet highly effective task: separating the entire moving foreground from the background. This yields high-quality masks for complex non-rigid objects, complementing the instance-level output of the first module. To this end, we employ a U-Net architecture~\cite{ronneberger2015unet} that takes bidirectional optical flow as input and outputs a single binary mask delineating all moving regions within the central frame.

\subsection{Mask Quality Assessment and Filtering}
\label{sec:filtering}
ROSM and MFSM produce a large volume of candidate masks, many of which are low-quality and can degrade the performance of downstream models. To prune these, we introduce a \textbf{Mask Quality Score} ($S_{\text{quality}}$), which combines two components to retain high-fidelity predictions:

\paragraph{Maskness ($S_{\text{maskness}}$)}
This score measures the model's confidence in the pixels it predicts as belonging to the mask. Given a soft mask prediction $p \in [0,1]^{H \times W}$, we define the set of positive pixels as $\mathcal{P}_{\text{pos}} = \{ (x,y) \mid p(x,y) > \tau_c \}$, where $\tau_c$ is a confidence threshold. The score is then the mean prediction value for this set of pixels:
\begin{equation}
    S_{\text{maskness}} = \frac{1}{|\mathcal{P}_{\text{pos}}|} \sum_{(x,y) \in \mathcal{P}_{\text{pos}}} p(x,y).
    \label{eq:maskness}
\end{equation}

\paragraph{Boundary Sharpness ($S_{\text{sharpness}}$)}
This score rewards masks with well-defined, sharp edges. Let $M_b$ be the mask binarized from $p$ at threshold $\tau_c$. Let $B$ be a $d$-pixel wide band around the contour of $M_b$, and $\nabla p$ be the gradient magnitude of $p$. The sharpness is the proportion of boundary pixels with a gradient magnitude above a threshold $\gamma_g$:
\begin{equation}
    S_{\text{sharpness}} = \frac{1}{|B|} \sum_{(x,y) \in B} \mathbb{I}(\|\nabla p(x,y)\| > \gamma_g),
    \label{eq:sharpness}
\end{equation}
where $\mathbb{I}(\cdot)$ is the indicator function.

The final quality score is a weighted sum of these two components: $S_{\text{quality}} = \beta \cdot S_{\text{maskness}} + (1 - \beta) \cdot S_{\text{sharpness}}$, where $\beta$ is a weighting factor. We discard masks with a score below a predefined threshold. Finally, we apply Non-Maximum Suppression (NMS)~\cite{neubeck2006efficient} to the filtered masks to eliminate duplicates, yielding the final set of pseudo-labels.

\subsection{Pseudo-Label Dataset Construction}
\label{ssec:dataset}

We applied our MGMS pipeline to a diverse collection of large-scale public video datasets\footnote{We emphasize that no manual annotations (e.g., labels, boxes, masks) from these datasets were used, and we ensured no overlap with our downstream evaluation benchmarks.}. The videos span a wide range of scenes, including human activities, animal behaviors, and driving footage. The complementary nature of our two segmentation modules allows us to generate a rich, multi-granularity dataset. The rigid instance module identifies distinct objects and parts, while the motion foreground module captures complete non-rigid entities. From approximately 10,000 hours of raw video, we generated around \textbf{21 million} high-quality motion pseudo-labels. This large-scale, automatically curated dataset forms the foundation for training our Perceptual Grouping Model, as detailed in the next section.

\section{Learning from Motion to Segment Anything}
\label{sec:Segment_Anything}

While the pseudo-labels from Section~\ref{sec:Pseudo_Labels_Generation} are derived from motion, our ultimate goal is to train a universal segmentation model that operates on static images, without any reliance on motion cues at inference time. However, a fundamental challenge arises from the nature of our pseudo-supervision: the masks generated from motion are high-quality but inherently sparse, covering only a fraction of the objects present in an image, e.g., one of the tiger's paws in Fig.~\ref{fig:stage_1}. To bridge this gap, we introduce the \textbf{P}erceptual \textbf{G}rouping \textbf{M}odel (\textbf{PGM}), a framework designed to learn a generalizable concept of ``objectness'' from this limited supervision.
Central to our approach is a novel contrastive training strategy, which enables the model to discover unlabeled objects beyond the provided masks.

\subsection{Perceptual Grouping Model Architecture}
PGM is built upon a standard Vision Transformer (ViT)~\cite{dosovitskiy2020image} backbone. To capture objects at various scales, PGM incorporates $K$ parallel linear projection heads on top of the ViT features. During training, each pseudo-mask is assigned to a specific head based on its bounding box size. The loss for that mask is computed exclusively using the features from its assigned head. This strategy encourages different heads to specialize in segmenting objects of different scales. Furthermore, to enhance robustness against occlusions and complex scenes, we employ the Copy-Paste data augmentation technique~\cite{ghiasi2021simple}.

\subsection{Perceptual Grouping Contrastive Learning (PGCL)}
Standard segmentation objectives, such as Binary Cross-Entropy (BCE)~\cite{ronneberger2015unet} or Dice loss~\cite{DBLP:conf/miccai/SudreLVOC17}, are suboptimal for our training data. Since motion-based pseudo-labels are inherently sparse, these losses erroneously penalize unannotated object instances as background. While Slot Attention~\cite{kara2024diod} addresses sparsity, it struggles in open-world settings: its fixed slot capacity clashes with the varying number of objects in open-world scenes, and its rigid pixel-to-slot assignment hampers multi-granularity segmentation. Crucially, slot-centric representations tend to overfit to the specific object categories seen during training, limiting \textit{zero-shot generalization} to unseen domains.

\begin{figure}[t]
    \centering
    \includegraphics[width=0.8\textwidth]{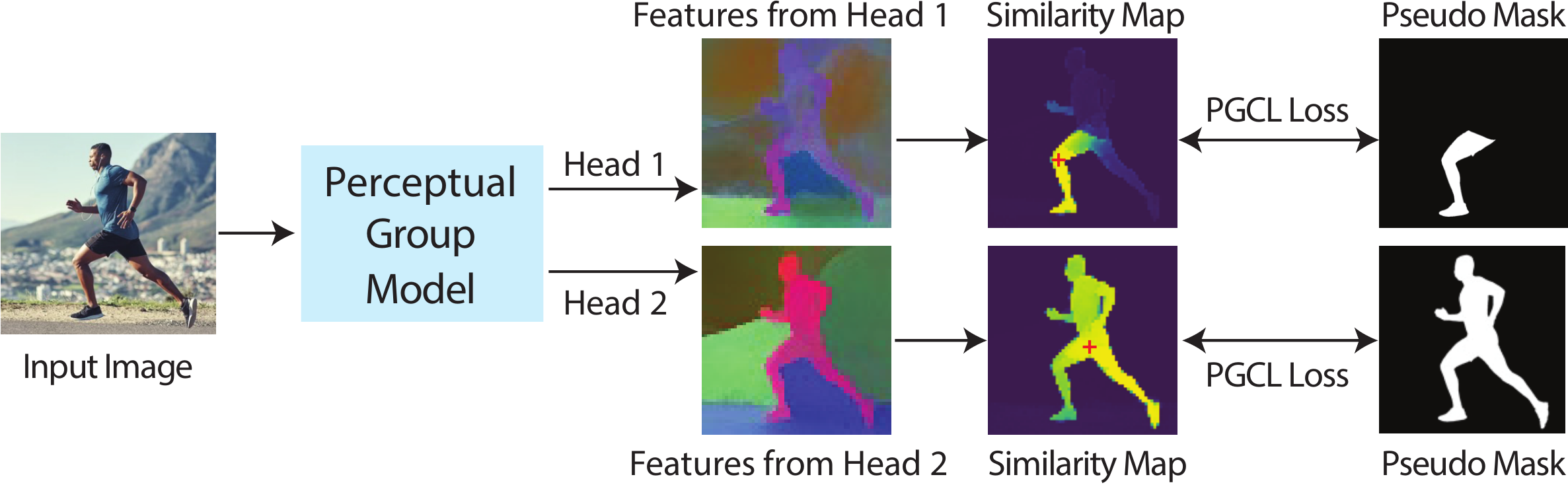}
    \caption{Illustration of Perceptual Grouping Contrastive Learning (PGCL). Our PGM features multiple heads, each trained on pseudo-masks of a specific scale range. For each assigned mask, we compute a similarity map by comparing an in-mask anchor patch (\textbf{+}) with all patches in the corresponding head's feature space. The PGCL loss supervises this map using the mask as the ground truth, training the model to group coherent patches into objects.}
    \label{image-stage-2}
\end{figure}

To overcome these significant limitations and cultivate a more robust and generalizable notion of ``objectness'' independent of explicit object count or categories, we introduce \textbf{Perceptual Grouping Contrastive Learning (PGCL)}, as illustrated in Fig.~\ref{image-stage-2}. The central idea is to impose intra-object feature compactness while promoting inter-object feature separability in the patch-level feature space. Specifically, for an input image tokenized into $N$ patches, each of the $K$ heads produces a set of patch embeddings $F^k \in \mathbb{R}^{N \times D}$. Consider a pseudo-mask $\mathcal{M}_i$ assigned to head $H_k$. We sample an anchor patch embedding $\mathbf{f}_t^k$ from within the mask region. We then compute its cosine similarity $s_{t,j}^k$ with all patch embeddings $\{\mathbf{f}_j^k\}_{j=1}^N$ from the same head. The probability $p_{t,j}^k$ that patch $j$ belongs to the same object as anchor $t$ is modeled via a sigmoid function with a learnable temperature $\tau_k$:
\begin{equation}
    p_{t,j}^{k}=\sigma(\tau_{k}\cdot s_{t,j}^{k}),
\end{equation}
where $\sigma(\cdot)$ is the sigmoid function.
We define a binary label $m_{t,j}$, which is 1 if both the anchor patch $t$ and patch $j$ are within the mask $\mathcal{M}_i$, and 0 otherwise. The PGCL loss is a BCE loss, averaged across all heads and sampled anchors:
\begin{equation}
\label{eq:pgcl}
    \mathcal{L}_{\text{PGCL}} = -\frac{1}{K N_a N} \sum_{k=1}^{K} \sum_{t=1}^{N_a} \sum_{j=1}^N \Big[  m_{t,j} \log p_{t,j}^k + (1 - m_{t,j}) \log (1 - p_{t,j}^k) \Big],
\end{equation}
where $N_a$ is the number of sampled anchors. By explicitly contrasting positive pairs (patches within the same mask) against negative ones, PGCL enables the model to capture object coherence based on feature similarity. This allows the network to generalize beyond the sparse pseudo-labels, effectively handling the ``segment anything'' task for unseen objects.

\subsection{Adaptation to Segment Anything}
\label{sec:adaptation_and_training}

While PGM learns a general notion of objectness, it is trained on relatively low-resolution video frames and lacks prompt-based interaction capabilities. To bridge this gap and enable high-performance, high-resolution segmentation with prompt-based interaction, we leverage PGM to generate high-quality masks on high-resolution static images. We then utilize these perceptual grouping masks to train two specialized models for distinct segment anything tasks.

\subsubsection{PGM Inference}
Our perceptual grouping masks generation begins by feeding a static image into the pre-trained PGM. PGM processes the image to generate an initial pool of candidate soft masks, derived from patch embeddings through pairwise cosine similarity. These candidates are then filtered based on a predicted mask quality score (Sec.~\ref{sec:filtering}) and refined using a dense Conditional Random Field (CRF)~\cite{lafferty2001conditional} to sharpen object boundaries. For high-resolution imagery, we employ a multi-scale, sliding-window approach: PGM inference is applied independently to overlapping tiles at various resolutions. Masks from all tiles are subsequently projected back, merged, and filtered via Non-Maximum Suppression (NMS) to remove duplicates and produce the final comprehensive and fine-grained results. Further implementation details can be found in Appendix~\ref{app:pgm_inference}.

\subsubsection{Training the Segment Anything Models}
We apply the aforementioned inference pipeline to generate perceptual grouping masks on a 1\% subset of the high-resolution SA-1B dataset~\cite{SAM}. We then train two Segment Anything models, following the approach of UnSAM~\cite{UnSAM}:

\begin{itemize}
    \item \textbf{Whole-Image Segmentation:} For universal object segmentation, we train a Mask2Former model~\cite{cheng2022masked}. This model accepts a full high-resolution image as input and automatically outputs a set of masks, where each mask corresponds to a distinct instance discovered in the scene.
    \item \textbf{Promptable Segmentation:} To build a prompt-driven model, we adopt an architecture based on Semantic-SAM~\cite{Semantic-SAM}, which excels at predicting masks at multiple granularity levels from a single click.
    To enable interactivity, we simulate user clicks during training by randomly sampling a point within the foreground of each mask to serve as a positive point prompt.
\end{itemize}

\section{Experiments}
\label{headings}

\subsection{Training Setups}
\subsubsection{MGMS Pipeline Training Settings}
Our MGMS Pipeline comprises two key modules: ROSM employs a SOLO-v2 architecture~\cite{wang2020solov2} with a ResNet-18 backbone~\cite{ResNet}, while MFSM utilizes a Res-UNet architecture~\cite{Res-UNet}, also with a ResNet-18 backbone. Both modules take a sequence of 7 frames as input and were trained on a synthetic dataset for 50,000 iterations with an input resolution of $512 \times 512$ and a batch size of 32.

\subsubsection{PGM Training Settings}
The training data for PGM comprises 21 million pseudo-labels distributed across 10 million static frames. This dataset was generated from a 10,000-hour video corpus constructed from three major sources: Kinetics-700~\cite{carreira2019short}, which contains approximately 650,000 10-second clips covering a wide range of human actions; BDD100K~\cite{yu2020bdd100k}, a large-scale driving video dataset designed for diverse road scenarios; and a 5\% random sample from YouTube-8M~\cite{abu2016youtube}, a broad-coverage video collection sourced from YouTube. PGM utilizes a ViT-base architecture~\cite{dosovitskiy2020image} with an input size of $512 \times 512$ pixels. The architecture is topped by four parallel heads, each implemented as a linear layer that produces 128-dimensional feature vectors. We employ the AdamW~\cite{loshchilov2017decoupled} optimizer with a learning rate of $1 \times 10^{-4}$ and a batch size of 32.

\subsubsection{Segment Anything Model Training Settings}
For whole-image segmentation, we construct a model by integrating a Mask2former~\cite{cheng2022masked} decoder with a ResNet-50~\cite{ResNet} backbone. Training is conducted for 8 epochs, configured with a $5 \times 10^{-5}$ learning rate, a batch size of 16, and 0.05 weight decay. For promptable segmentation, we leverage the Semantic-SAM~\cite{Semantic-SAM} architecture with a Swin-Transformer~\cite{Swin} Tiny model as the backbone to improve performance. Both models are trained with only a 1\% subset of unlabeled images from SA-1B~\cite{SAM}.

\begin{table*}[t]
    \begin{center}
    \caption{Evaluation results of PGM on seven datasets. $\textsuperscript{\dag}$: The pseudo-labels used by UnSAM~\cite{UnSAM} are not strictly unsupervised as they are refined by CascadePSP~\cite{CascadePSP}.}
    \label{table_Pseudo-Masks}
    \begin{tabular}{l|ccccccc|c}
    \toprule
    Methods & PtIn & LVIS & Entity & PACO & ADE & COCO & SA-1B & \textbf{Average} \\
    \midrule
    UnSAM~\cite{UnSAM} (CRF~\cite{lafferty2001conditional})& 23.2 & 16.5 & 22.1 & 10.8 & 15.1 & 21.9 & 15.3 & 17.8 \\
    UnSAM~\cite{UnSAM} (CascadePSP$\textsuperscript{\dag}$~\cite{CascadePSP})& 27.3 & 19.9 & 24.3 & 12.8 & 17.8 & 24.8 & 24.7 & 21.7 \\
    \midrule
    PGM~(\textbf{ours}) (w/o refinement) & 34.7 & 27.9 & 28.2 & 16.1 & 25.5 & 32.6 & 36.6 & 28.8 \\
    PGM~(\textbf{ours}) (CRF~\cite{lafferty2001conditional}) & \textbf{36.1} & \textbf{29.3} & \textbf{30.1} & \textbf{16.9} & \textbf{26.7} & \textbf{33.8} & \textbf{37.8} & \textbf{30.1} \\
    \bottomrule
    \end{tabular}
    \end{center}
\end{table*}

\subsection{Evaluation Datasets and Metrics}

We evaluate MoSA under two segmentation settings: \textit{Whole-Image Segmentation} and \textit{Point-Based Promptable Segmentation}. For Whole-Image Segmentation, we evaluate MoSA on seven benchmark datasets, including COCO~\cite{lin2014microsoft}, LVIS~\cite{Gupta_2019_CVPR}, ADE20K~\cite{zhou2019semantic}, EntitySeg~\cite{qi2022open}, SA-1B~\cite{SAM}, PartImageNet~\cite{he2022partimagenet}, and PACO~\cite{ramanathan2023paco}. Importantly, each dataset only labels a subset of objects, whereas our model generates masks across heterogeneous levels and open-world categories. As a result, the standard COCO-style Average Precision (AP) metric may not fully capture the model's ability to segment diverse, open-world entities. Therefore, following prior unsupervised segmentation works~\cite{CutLER, UnSAM, cao2024sohes}, we adopt Average Recall (AR) as the primary metric for whole-image segmentation comparisons. For Point-Based Promptable Segmentation, we evaluate on COCO Val2017~\cite{lin2014microsoft}. Following prior promptable segmentation methods~\cite{SAM,Semantic-SAM}, we report two metrics: MaxIoU and OracleIoU. MaxIoU measures the Intersection over Union (IoU) between the ground truth mask and the highest-confidence predicted mask, while OracleIoU reports the maximum IoU achieved among all predicted masks.

\begin{figure}[t]
    \centering
    \includegraphics[width=0.8\textwidth]{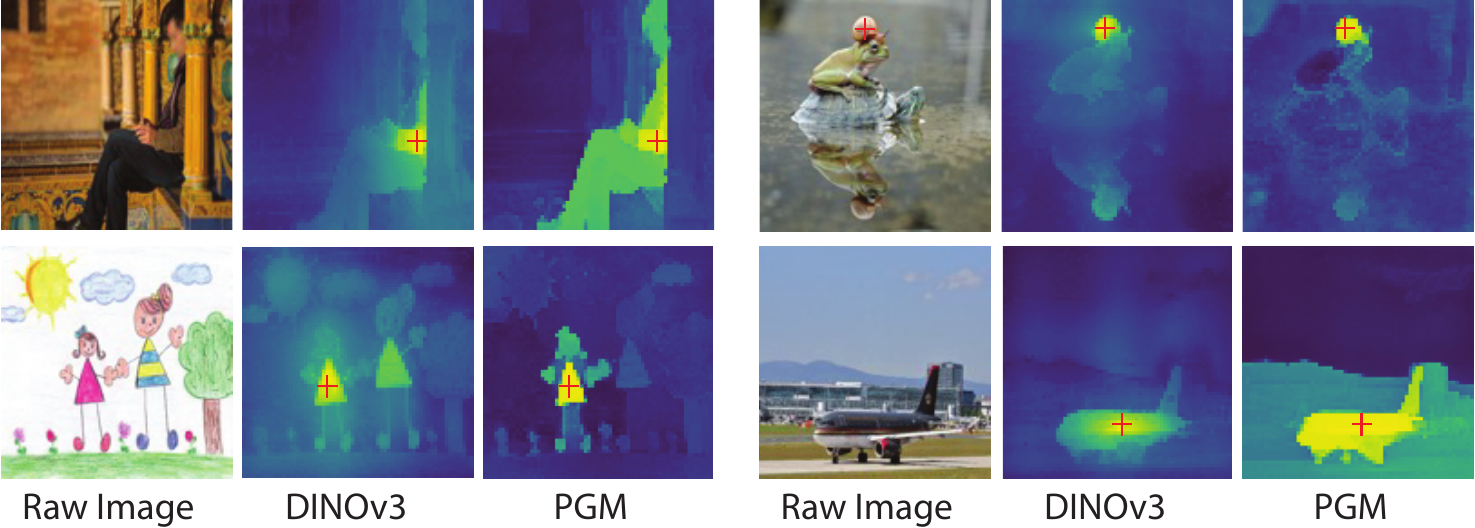}
    \caption{Qualitative comparison of PGM (ours) with DINOv3 family~\cite{DINOv3}, where the red \textbf{+} denotes anchor patches.}
    \label{image-pgm-vs-dino}
\end{figure}

\subsection{Evaluation Results}
\subsubsection{Perceptual Grouping Masks}
We quantitatively evaluate PGM on seven benchmarks. As shown in Table~\ref{table_Pseudo-Masks}, PGM generates high-quality masks. Notably, even without refinement, they significantly outperform the pseudo-labels used by UnSAM~\cite{UnSAM}, which are post-processed with CascadePSP\cite{CascadePSP} (a method relying on manual annotations). Qualitatively, Fig.~\ref{image-pgm-vs-dino} compares feature similarity maps of PGM and the strong self-supervised vision model DINOv3~\cite{DINOv3}. The results highlight PGM's superior ability to learn a holistic concept of ``objectness''. For instance, in the top-right image, DINOv3 only activates on part of the snail, whereas PGM correctly segments the entire snail. Moreover, PGM shows remarkable zero-shot generalization. When applied to an out-of-distribution hand-drawn image (bottom-left), it successfully identifies the child \mbox{figure}, whereas the baseline fails to grasp this abstract concept. This suggests that our PGCL effectively teaches the model a generalizable notion of object coherence from sparse motion-based supervision.

\begin{table*}[ht]
    \centering
    \caption{Results on the whole-image segmentation task. $\textsuperscript{\dag}$: UnSAM~\cite{UnSAM} utilizes CascadePSP~\cite{CascadePSP}, a method that relies on manual annotations, and is therefore not strictly unsupervised. MoSA achieves SOTA performance among unsupervised methods only using a ResNet-50 backbone and delivers results comparable to SAM~\cite{SAM}.}
    \label{tab_Whole-Image}
    \begin{tabular}{llc|cccccc|cc}
    \hline
    \multirow{2}{*}{Methods} & Backbone & \#  & \multirow{2}{*}{\textbf{Avg.}}   & \multicolumn{5}{c|}{with Whole Entities}  & \multicolumn{2}{c}{ with Parts}\\
    \cline{5-9}   \cline{10-11}
                                       &  \#params   & images &  & COCO &LVIS &ADE  &Entity &SA-1B &PtIn &PACO\\
        \hline
        \textcolor{gray}{SAM~\cite{SAM}}       &\textcolor{gray}{VIT-B(85M)}  &\textcolor{gray}{11M}  &\textcolor{gray}{42.1} &\textcolor{gray}{49.6} &\textcolor{gray}{46.1} &\textcolor{gray}{45.8} &\textcolor{gray}{45.9} &\textcolor{gray}{60.8} &\textcolor{gray}{28.3} &\textcolor{gray}{18.1}\\
        \hline
        {FreeSOLO~\cite{wang2022freesolo}}   &RN-101 (45M) & 1.3M & 7.3 & 11.6 & 5.9 & 7.3 & 8.0 & 2.2 & 13.8 & 2.4 \\
        {CutLER~\cite{CutLER}} & RN-50 (23M) & 1.3M & 21.8 & 28.1 & 20.2 & 26.3 & 23.1 & 17.0 & 28.7 & 8.9\\
        {SOHES~\cite{cao2024sohes}}             &VIT-B(85M)  &0.2M  &30.1 &30.5 &29.1 &31.1 &33.5 &33.3 &36.0 &17.1\\
        {UnSAM$\textsuperscript{\dag}$~\cite{UnSAM}}       &RN-50(23M)  &0.1M  &39.2 &40.5 &37.7 &35.7 &39.6 &41.9 &51.6 &27.5\\
        \hline
        {MoSA~(\textbf{ours})} &RN-50(23M)  &0.1M  &\textbf{42.1} &\textbf{43.5} &\textbf{42.2} &\textbf{38.4} &\textbf{41.1} &\textbf{48.2} &\textbf{52.7} &\textbf{28.4}\\
        \hline
    \end{tabular}
\end{table*}

\subsubsection{Whole-Image Segmentation}
Table \ref{tab_Whole-Image} presents the whole-image segmentation performance, measured by AR, of our unsupervised MoSA against baselines. Notably, MoSA, trained with only 1\% of the SA-1B training data and utilizing a ResNet-50 backbone (only 23M parameters), achieves an average AR of 42.1\%. This performance significantly outperforms previous unsupervised methods and the improvement is consistent across all seven datasets. Furthermore, on PartImageNet~\cite{he2022partimagenet} and PACO~\cite{ramanathan2023paco}, MoSA surpasses the supervised SAM (ViT-B, 11M images) by 24.4\% and 10.3\% in AR, respectively. This suggests that pre-training on video motion cues enables the discovery of fine-grained objects and details often overlooked by human annotators. Qualitative results in Fig.~\ref{fig:title_image} further illustrate this; notably, MoSA effectively segments even completely static objects, such as tiles on a roof. This capability is a direct consequence of PGCL, which extrapolates from sparse, motion-derived pseudo-labels to learn a universal concept of ``objectness''. MoSA's results align more closely with human perception of object instances and exhibit less segmentation noise.

\subsubsection{Point-Based Promptable Segmentation}
As shown in Table~\ref{table_Point-Based}, MoSA achieves 41.6\% MaxIoU and 63.4\% OracleIoU on COCO~\cite{lin2014microsoft}. This notably outperforms the previous SOTA, UnSAM~\cite{UnSAM}, by +1.3\% and +3.9\% respectively. Crucially, UnSAM is not strictly unsupervised, as it relies on CascadePSP~\cite{CascadePSP}, a refiner trained with manual annotations. While fully supervised SAM~\cite{SAM} sets a higher benchmark (52.1\% MaxIoU, 68.2\% OracleIoU), MoSA's performance is compelling given its unsupervised nature and high efficiency, requiring 3.4$\times$ fewer parameters and 100$\times$ less SA-1B data. Qualitatively (Fig.~\ref{fig:title_image}), MoSA provides stable, perceptually-aligned, multi-granularity labels. For instance, even in challenging scenarios such as segmenting a person riding a horse (where their motion is often congruent), MoSA leverages video-learned semantics to accurately distinguish them. Additional qualitative results are in Appendix~\ref{app:qualitative_results}.

\begin{table}[!t]
    \centering
    \caption{Quantitative comparison on the point-based promptable image segmentation task. MoSA achieves superior performance over the previous SOTA, UnSAM~\cite{UnSAM}, which is not strictly unsupervised as it employs CascadePSP~\cite{CascadePSP}.}
    \label{table_Point-Based}
    \begin{tabular}{llc|cc}
    \toprule
            \multirow{2}{*}{Methods} & Backbone & \multirow{2}{*}{\% of SA-1B} & Point & Point  \\
            & (\# params) & &(Max) & (Oracle)   \\
        \midrule
        \textcolor{gray}{SAM~\cite{SAM} (supervised)} & \textcolor{gray}{ViT-B/8 (85M)} & \textcolor{gray}{100\%} & \textcolor{gray}{52.1} & \textcolor{gray}{68.2} \\
        \midrule
        UnSAM~\cite{UnSAM} & Swin-Tiny (25M) &1\% & 40.3 & 59.5 \\
        MoSA~(\textbf{ours}) & Swin-Tiny (25M) & 1\% & \textbf{41.6} & \textbf{63.4}\\
        \bottomrule
    \end{tabular}
\end{table}

\subsection{Ablation Study}

\subsubsection{PGM Training}
To verify the effectiveness of our proposed PGCL, we conduct a comparative analysis against strong baselines using identical backbone architectures and training data (1,000 hours of video). Specifically, we compare PGCL with: (1) standard \textbf{BCE+Dice} loss, and (2) \textbf{Slot Attention}~\cite{kara2024diod} with 256 slots. We evaluate the zero-shot transfer performance on 1,000 images from the SA-1B dataset, reporting Average Recall (AR) overall (AR$_{\text{1000}}$) and across varying scales (AR$_{\text{s}}$, AR$_{\text{m}}$, AR$_{\text{l}}$). As presented in Table~\ref{tab:pgm_ablation}, the conventional BCE+Dice baseline fails to segment most objects because sparse motion pseudo-labels wrongly suppress unlabeled static regions as background. Similarly, Slot Attention exhibits suboptimal performance (8.2 AR). We attribute this to the limitation of fixed slot capacity, which struggles to accommodate the uncertain number of objects and the multi-granular hierarchies inherent in open-world scenes. In contrast, our method achieves a substantial improvement, reaching 28.3 AR. This result confirms the superiority of PGCL in handling the complexity of the ``segment anything'' task.

\begin{table}[t] %
    \centering
    \caption{Ablation study of PGM training.}
    \label{tab:pgm_ablation}
    \begin{tabular}{l c c c c }
        \toprule
        Method & AR$_{\text{1000}}$ & AR$_{\text{s}}$ & AR$_{\text{m}}$ & AR$_{\text{l}}$ \\
        \midrule
        BCE+Dice & 3.8 & 1.2 & 2.8 & 6.6 \\
        Slot Attention~\cite{kara2024diod}  & 8.2 & 5.2 & 8.6 & 8.7 \\
        PGCL (ours) & \textbf{28.3} & \textbf{14.7} & \textbf{26.6} & \textbf{36.1} \\
        \bottomrule
    \end{tabular}
\end{table}

\subsubsection{MGMS Pipeline}
We ablate the MGMS Pipeline's key components: the Rigid Object (ROSM) and Motion Foreground (MFSM) Segmentation Modules. Their efficacy is evaluated based on PGM's Average Recall (AR) on 1,000 SA-1B images when trained on their generated pseudo-labels. As shown in Table~\ref{tab:ablation_mgms}, ROSM alone excels on small objects (20.6\% AR$_{\text{s}}$) but struggles with larger ones. Conversely, MFSM is effective for large objects (43.1\% AR$_{\text{l}}$) but performs poorly on small ones (13.7\% AR$_{\text{s}}$). The full pipeline integrates both modules to achieve the best overall score (36.6\% AR$_{\text{1000}}$). This confirms their complementary nature: combining ROSM's precision for small objects with MFSM's broader motion capture is essential for generating robust, multi-scale motion cues.

\begin{table}[t] %
  \centering
  \caption{Ablation study of modules in the MGMS pipeline.}
  \label{tab:ablation_mgms}
  \begin{tabular}{cc|cccc} %
    \toprule
    ROSM & MFSM  & AR$_\text{1000}$ &  AR$_\text{s}$ &  AR$_\text{m}$ &  AR$_\text{l}$ \\
    \midrule
    \checkmark &  & 28.8 & 20.6 & 31.1 & 28.7 \\
     & \checkmark & 34.7 & 13.7 & 34.9 & 43.1 \\
    \checkmark & \checkmark & \textbf{36.6} & \textbf{20.6} & \textbf{35.4} & \textbf{45.1} \\
    \bottomrule
  \end{tabular}
\end{table}

\subsubsection{Number of Prediction Heads}
We ablate the number of prediction heads in PGM for multi-granularity segmentation. As shown in Fig.~\ref{fig_ablations} (left), performance improves substantially from 27.9\% AR with a single head to 36.6\% AR with four heads. The most significant gain occurs when adding the second head (+4.2\% AR), confirming the fundamental benefit of specialized heads. However, performance saturates beyond four heads, dropping marginally to 36.5\% AR with five. This indicates that four heads provide the optimal trade-off, effectively capturing the full range of object scales from motion cues without introducing redundancy.

\subsubsection{Impact of Pre-training Video Data Scale}
\begin{figure}[t]
    \centering
    \includegraphics[page=3,width=0.7\textwidth]{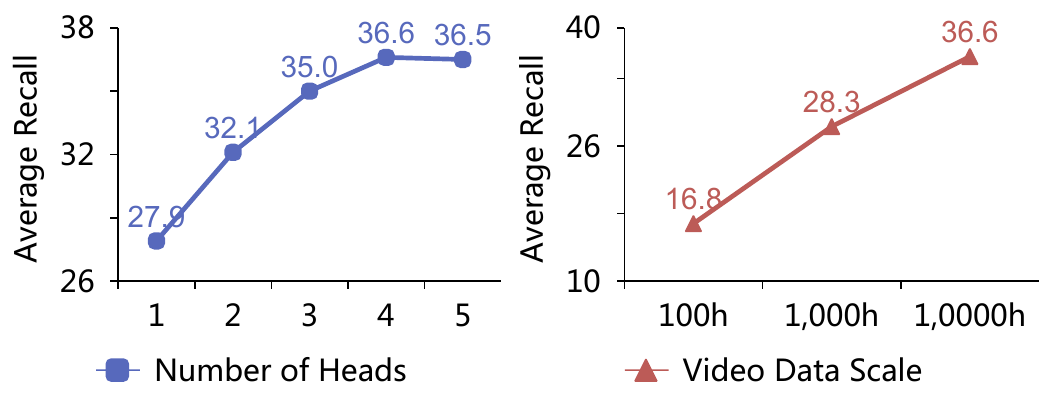}
    \caption{Impact of the number of prediction heads (left) and pre-training video data scale (right) in PGM.}
    \label{fig_ablations}
\end{figure}

We investigate how the amount of unlabeled video data impacts PGM's ability to learn motion priors. As illustrated on the right side of Fig.~\ref{fig_ablations}, we observe a strong positive correlation between data scale and performance. Specifically, increasing the training data from 100h to 1,000h, and further to 10,000h, substantially boosts AR from 16.8\% to 28.3\% and finally to 36.6\%. This clear, non-saturating trend underscores that our PGM effectively leverages large-scale video corpora and strongly suggests that its performance can be further enhanced by scaling to even larger datasets.

\section{Conclusion}
\label{conclusion}

Perceptual grouping is fundamental to human visual intelligence, enabling us to transform raw visual data into semantic understanding through simple, yet powerful rules. Inspired by this biological process, we introduce MoSA, a novel three-stage unsupervised framework for learning generalized segmentation from motion. Extensive experiments demonstrate that by observing visual motion alone, MoSA learns robust segmentation abilities applicable to diverse objects and scenes. This approach not only outperforms existing unsupervised methods, but also offers a more interpretable and scalable path toward realizing ``segment-anything.'' Our work highlights the immense potential of learning object perception from motion as a fundamental building block for artificial visual intelligence, bridging the gap towards human-like perceptual understanding.

\clearpage
\bibliographystyle{splncs04}
\bibliography{main}

\clearpage
\appendix
\section*{Appendix}
\setcounter{figure}{0}
\setcounter{table}{0}
\setcounter{equation}{0}
\renewcommand{\thefigure}{A\arabic{figure}}
\renewcommand{\thetable}{A\arabic{table}}
\renewcommand{\theequation}{A\arabic{equation}}
\renewcommand{\theHfigure}{appendix.\arabic{figure}}
\renewcommand{\theHtable}{appendix.\arabic{table}}
\renewcommand{\theHequation}{appendix.\arabic{equation}}
\section{Synthetic Video Generation Details}
\label{app:synthetic_generation}
Each synthetic video is generated by first rendering a dynamic background and then superimposing multiple moving foreground objects. The key components of this process are detailed below. Notably, the entire synthesis process is unsupervised and does not rely on manually annotated data; all ground-truth masks are generated programmatically.

\subsection{Background}
The dynamic background is designed to simulate various camera movements and global scene changes.

\subsubsection{Source} For each synthetic video, we randomly select a single image from a data source that is distinct from all evaluation benchmarks to serve as the base background texture.

\subsubsection{Motion Generation}
The static background image is animated by moving it along a smooth, randomly generated trajectory. This trajectory is defined by a sequence of 2D control points interpolated with B-splines. Concurrently, the background undergoes a random perspective transformation between frames. The parameters for these transformations are sampled randomly but are constrained to ensure smooth, temporally coherent changes, mimicking realistic camera motion.

\subsubsection{Motion Profile} To introduce variability in relative motion speeds, we simulate two primary scenarios:
\begin{itemize}
    \item \textbf{Fast Camera Motion}: The background undergoes rapid transformations, such as large inter-frame displacements and perspective shifts. In this scenario, foreground objects are set to move more slowly.
    \item \textbf{Slow Camera Motion}: The background exhibits slower, more subtle movements. Correspondingly, foreground objects exhibit faster and more pronounced motion.
\end{itemize}

\subsection{Foreground Objects}
Foreground objects are synthesized to introduce diverse shapes, appearances, and independent motion patterns, which are critical for learning motion-based segmentation.

\subsubsection{Source}
Object shapes are randomly generated in accordance with the procedure in OCLR~\cite{OCLR}, with textures sourced from a data collection kept separate from all evaluation benchmarks.

\subsubsection{Appearance Generation}
The visual appearance of each foreground object is diversified through a two-step process.
First, the generated object mask undergoes random resizing and in-plane rotation, with angles sampled uniformly from $[0, 360^\circ)$.
Second, the transformed mask is treated as a sprite and re-textured. This involves randomly selecting an image, cropping a random patch from it, and applying this patch as the new texture for the sprite.

\subsubsection{Motion Generation}
To simulate dynamic scenes with varying complexity, the number of distinct moving foreground objects per video is randomly sampled from 1 to 15. Each object is initialized at a random 2D position in the first frame and follows an independently generated, smooth B-spline trajectory. Concurrent with its translation, each object also undergoes its own sequence of random perspective transformations across frames. These transformations are constrained to ensure inter-frame smoothness, simulating plausible object movement and deformation.

\subsubsection{Motion Disambiguation}
A critical aspect of our synthesis is ensuring that motion cues are sufficient to distinguish individual objects and separate them from the background, especially for optical flow-based models. To this end, we enforce two key constraints.
First, the motion trajectories (including translation, velocity, and perspective changes over time) of any two foreground objects within the same video are made unique by sampling different control points and parameters for their respective B-spline paths and transformations.
Second, the motion trajectory of each foreground object is also guaranteed to be distinct from the global background motion.

\begin{figure*}[t]
    \centering
    \includegraphics[width=\textwidth]{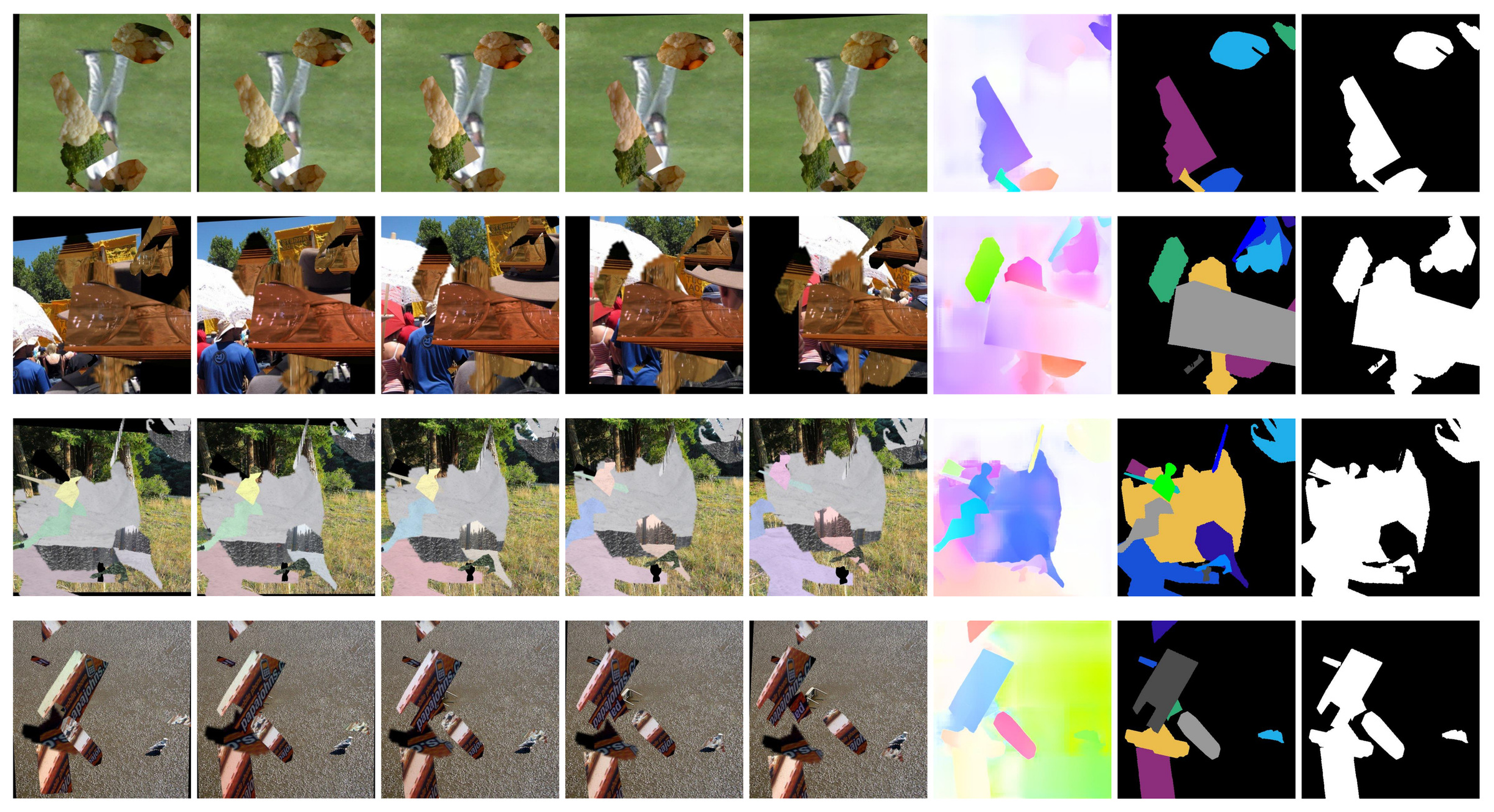}
    \caption{Examples from our synthetic video dataset, showcasing challenging scenarios with significant occlusions and complex object interactions. Each row presents one example. Columns 1-5: Five sequential video frames. Column 6: Optical flow for the central frame. Column 7: Individual instance masks for each moving object in the central frame. Column 8: The complete motion foreground segmentation mask for the central frame.}
    \label{fig_1}
\end{figure*}

\subsection{Synthetic Data Visualizations}
Fig.~\ref{fig_1} provides visual examples from our synthetic dataset. These examples are specifically designed to be challenging, featuring heavy inter-object occlusions and complex motion trajectories that emulate difficulties found in real-world scenes. The visualizations include the input frames, the resulting optical flow, and the ground-truth segmentation masks, illustrating the direct link between the generated motion and the desired output. It is important to emphasize that the visual fidelity of textures and the precise shapes of the foreground objects are not the primary goal of our synthesis. Since our downstream model relies solely on optical flow as input, it is invariant to such high-level semantic cues. The key objective of our data generation is to produce diverse and complex \textit{motion patterns}, which are effectively captured by the optical flow representation.

\section{Motion Pseudo-Labels Generation Details}
\label{sec:video_data_processing}

Serving as the critical data foundation of MoSA, the Multi-Granularity Motion Segmentation (MGMS) pipeline is designed to automatically generate high-quality motion pseudo-labels from large-scale unannotated videos. This section details how this framework processes unlabeled video data to extract robust motion cues. We first introduce the diverse video datasets that serve as input, followed by a comprehensive description of our preprocessing steps and the outputs derived from the specialized modules of the MGMS pipeline.

\subsection{Dataset Description}
\label{ssec:dataset_description}

The training data for our model are derived from three large-scale public video datasets:
\begin{itemize}
    \item \textbf{Kinetics-700~\cite{carreira2019short}}: This dataset consists of approximately 650,000 video clips, encompassing 700 distinct human action classes. The videos feature a wide array of human-object interactions (e.g., playing instruments) and human-human interactions (e.g., shaking hands, hugging). Each action class is represented by at least 700 video clips, with each clip having an approximate duration of 10 seconds.

    \item  \textbf{BDD100K~\cite{yu2020bdd100k}}: A substantial driving video dataset containing 100,000 videos. It is designed for evaluating image recognition algorithms in the context of autonomous driving across 10 different tasks. The dataset is characterized by its diversity in geographic locations, environmental conditions, and weather patterns.

    \item \textbf{YouTube-8M~\cite{abu2016youtube}}: A very large-scale video dataset comprising over 7 million videos, annotated with 4,716 classes by an automated system. Videos are categorized into 24 broad topics based on their visual content, including sports, gaming, arts \& entertainment, among others. For our experiments, we utilized a random sample of 5\% of the videos from this dataset.
\end{itemize}

It is important to reiterate that we did not utilize any human-provided annotations (such as labels, object bounding boxes, or segmentation masks) from these datasets. Our methodology relies solely on the visual content of the videos. Collectively, these datasets provide a rich and diverse collection of scenarios, encompassing human motion, object interactions, animal behavior, autonomous driving footage, and first-person perspectives. This diversity is crucial for training robust models capable of understanding complex dynamic scenes.

\subsection{Data Preprocessing}
\label{ssec:data_preprocessing}

Our data processing pipeline begins with an initial filtering step to ensure that we only process videos with significant motion. To achieve this, the raw videos from the aforementioned datasets were processed by sampling clips at fixed 5-second intervals.
We compute optical flow for sampled video clips using GMFlow. We then calculate the mean magnitude of the optical flow vectors across all pixels and frames. Clips where this mean magnitude falls below a predefined threshold are discarded, as they likely contain only static scenes or negligible camera jitter. This step focuses our computational resources on dynamically rich content.

For each remaining clip, we extract a central 1.5-second segment and uniformly sample 7 frames. We compute bidirectional optical flow between adjacent frames with GMFlow. The optical flow fields from the central 7 frames yield 24 input channels (6 forward and 6 backward flows, each with 2 channels). These are fed into two modules: the Rigid Object Segmentation Module (ROSM) and the Motion Foreground Segmentation Module (MFSM).

\subsection{Module Outputs}
\label{ssec:module_outputs}
\begin{figure*}[t]
    \centering
    \includegraphics[width=1\textwidth]{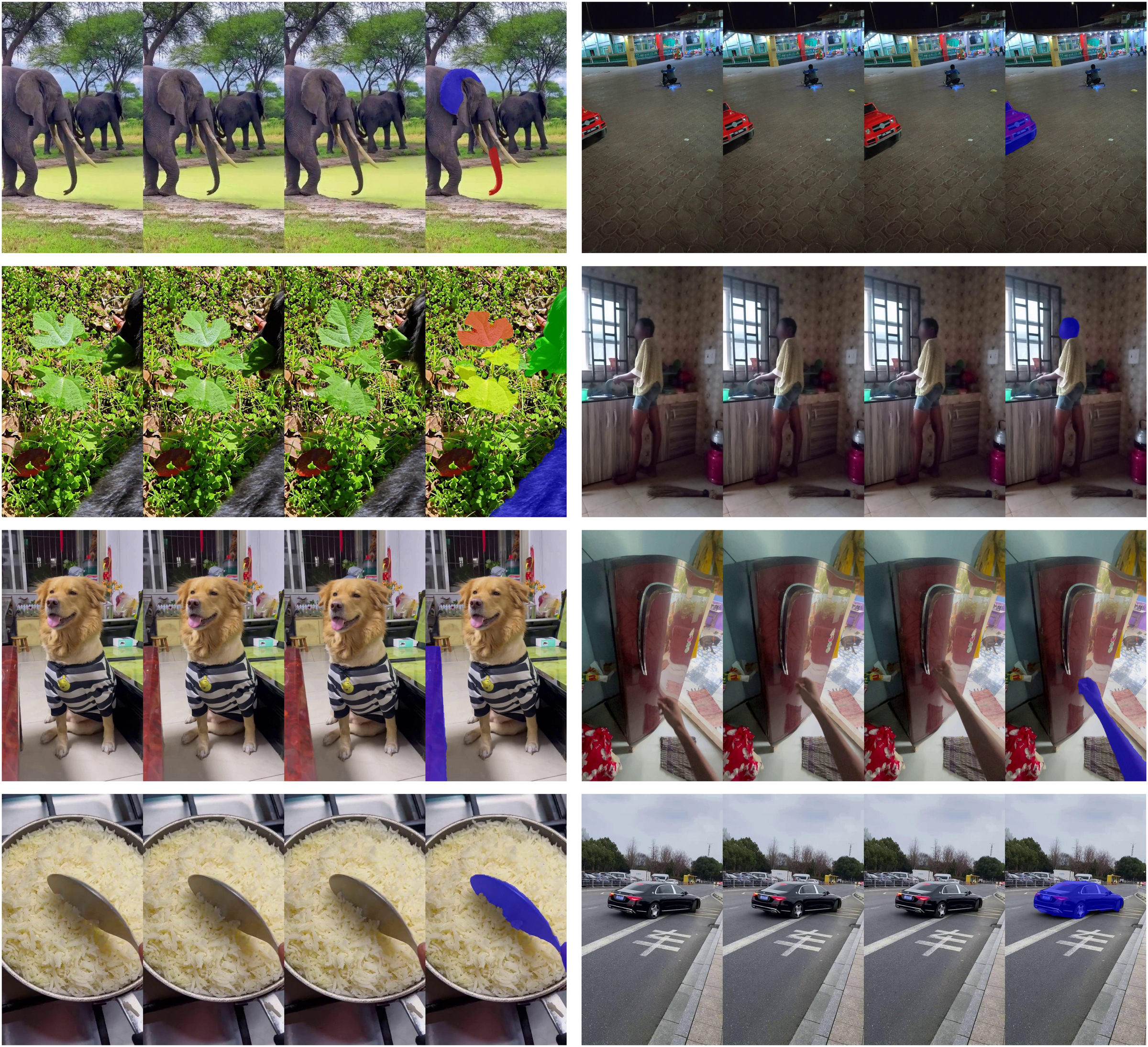}
    \caption{Rigid Object Segmentation Module (ROSM) output examples. For each case, the first three columns show key input frames, and the fourth column presents ROSM’s segmentation of the dominant rigidly moving entity. ROSM effectively segments whole rigid objects (cars, spoon), isolates articulated parts (elephant’s ear/trunk, human head), and distinguishes smaller moving components (leaves). Furthermore, ROSM can leverage parallax cues from camera motion to segment static objects(e.g.,  table next to the dog).}
    \label{fig:rosm_output}
\end{figure*}

\begin{figure*}[t]
    \centering
    \includegraphics[width=1\textwidth]{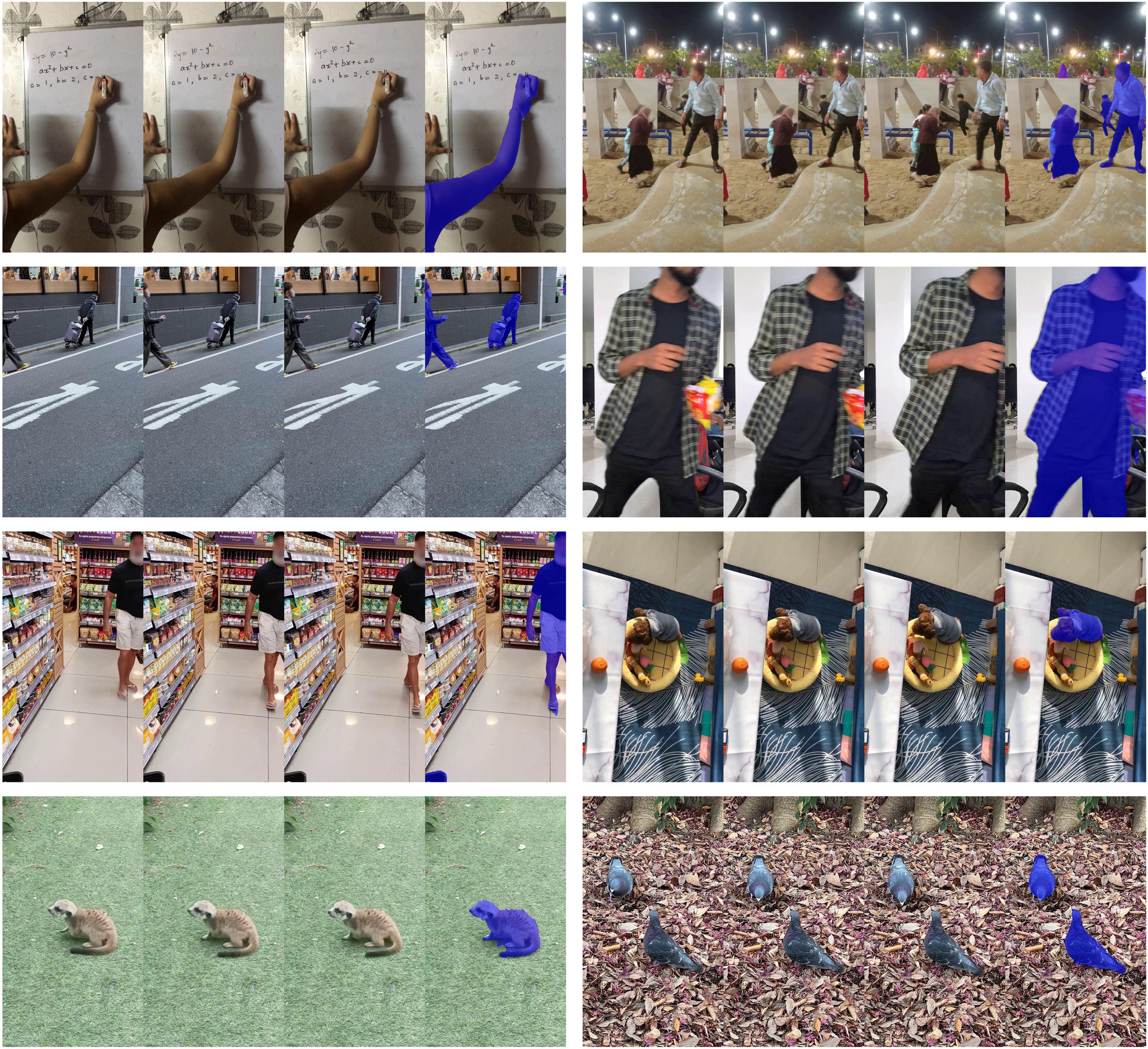}
    \caption{Motion Foreground Segmentation Module (MFSM) effectively extracts all moving foreground elements. For each case, the first three columns display representative input frames, while the fourth column shows MFSM's segmentation results.}
    \label{fig:mfsm_output}
\end{figure*}

The outputs of the ROSM and MFSM are illustrated in Fig.~\ref{fig:rosm_output} and Fig.~\ref{fig:mfsm_output}, respectively. Specifically, ROSM (Fig.~\ref{fig:rosm_output}) not only identifies rigid moving objects (e.g., spoons, cars) but also leverages parallax to segment static foreground objects (e.g., a table next to the dog) and discerns object parts (e.g., an elephant's trunk, a human arm). In contrast, MFSM (Fig.~\ref{fig:mfsm_output}) extracts a broader range of moving foreground elements, including pedestrians, animals, and human-object interactions.

\subsection{Data Postprocessing}
\label{ssec:data_postprocessing}
The raw masks generated by these two modules undergo a rigorous post-processing pipeline to ensure their quality and uniqueness.
First, to prune low-fidelity predictions, we introduce a \textbf{Mask Quality Score} ($S_{\text{quality}}$) for each candidate mask.
This score synergizes two critical metrics. The first, \emph{maskness} ($S_{\text{maskness}}$), quantifies model confidence by averaging the prediction values for all pixels exceeding a threshold of $\tau_c = 0.5$.
The second, \emph{boundary sharpness} ($S_{\text{sharpness}}$), evaluates contour clarity by calculating the proportion of high-gradient pixels (magnitude $>\gamma_g = 0.3$) within a 4-pixel-wide band around the mask's binarized edge.
The final score is a weighted sum: $S_{\text{quality}} = \beta \cdot S_{\text{maskness}} + (1 - \beta) \cdot S_{\text{sharpness}}$, where we set $\beta=0.4$ to place a slightly higher emphasis on boundary sharpness.
Masks with a score below a stringent quality threshold of $\tau_{\text{quality}}=0.85$ are discarded.
Subsequently, to eliminate redundant masks for the same object, we employ Non-Maximum Suppression (NMS)~\cite{neubeck2006efficient}.
Finally, since frames with significant motion can suffer from blur, we filter the generated image-label pairs based on image sharpness. We assess this by calculating the variance of the Laplacian of the image region defined by the mask, and discard pairs where this variance falls below a threshold $\tau_{\text{blur}}$.
This multi-stage filtering protocol is crucial for populating our final dataset with sharp, distinct, and high-confidence pseudo-labels.

\section{Model Implementation Details}
\label{sec:architecture_details}

To provide a comprehensive understanding of MoSA, this section outlines the architectural design and implementation specifics of its key components. We will first describe the \textbf{Multi-Granularity Motion Segmentation (MGMS)} pipeline, which extracts candidate object masks from unlabeled videos by identifying coherently moving regions. Subsequently, we will detail the \textbf{Perceptual Grouping Model (PGM)}, focusing on its Vision Transformer architecture~\cite{dosovitskiy2020image} and the training and inference procedures. Lastly, we will describe the \textbf{Whole-Image Segmentation} and \textbf{Promptable Segmentation} models. These models are designed to distill object concepts learned by the PGM, enabling high-resolution segmentation capabilities across entire images, akin to a ``segment anything'' functionality.

\begin{table*}[t]
  \centering
  \caption{Training Hyperparameters for ROSM and MFSM.}
  \label{tab:all_module_hyperparams}
  \begin{tabular}{l|ll}
    \toprule
    Config & ROSM & MFSM \\
    \midrule
    Optimizer & AdamW~\cite{loshchilov2017decoupled} & AdamW~\cite{loshchilov2017decoupled} \\
    Base learning rate & $5 \times 10^{-4}$ & $1 \times 10^{-4}$ \\
    Weight decay & $1 \times 10^{-4}$ & $1 \times 10^{-4}$ \\
    Optimizer momentum & $\beta_1=0.9, \beta_2=0.999$ & $\beta_1=0.9, \beta_2=0.999$ \\
    Batch size & 32 & 32 \\
    Learning rate schedule & Linear Decay & Linear Decay \\
    Warmup iterations & 1000 & 0 \\
    Number of iterations & 60000 & 30000 \\
    Augmentation & Random flip & Random flip \\
    Input frames & 7 & 7 \\
    Input resolution & $512 \times 512$ & $512 \times 512$ \\
    Input channels & 24 & 24 \\
    \bottomrule
  \end{tabular}
\end{table*}
  \begin{table*}[t!]
      \centering
      \small
      \caption{Training Hyperparameters for the Perceptual Grouping Model (PGM).}
      \label{tab:pgm_hyperparams}
      \begin{tabular}{l|l}
      \toprule
      Hyperparameter & Value \\
      \midrule
      Optimizer & AdamW~\cite{loshchilov2017decoupled} \\
      Base Learning Rate & 1e-4 \\
      Weight Decay & 1e-2 \\
      Optimizer Momentum & $\beta_1=0.9, \beta_2=0.999$ \\
      Batch Size & 32 \\
      Learning Rate Schedule & Linear Decay \\
      Warmup Epochs & 0.1 \\
      Total Epochs & 10 \\
      Data Augmentation & Random Flip, ColorJitter, Copy-Paste, Random Crop \\
      Number of Projection Heads & 4 \\
      Head Output Feature Dimension & 128 \\
      Pretrained Backbone & MAE ViT-B/8~\cite{DINO} \\
      Input Resolution & $512 \times 512$ \\
      Number of Anchor Patches & 6 \\
      \bottomrule
      \end{tabular}
  \end{table*}
\subsection{MGMS Pipeline}
\label{app:mgms_training}

Our MGMS Pipeline is composed of two interconnected modules: Rigid Object Segmentation Module (ROSM) and Motion Foreground Segmentation Module (MFSM). We now detail the model architecture and training procedures for each of these modules. Table~\ref{tab:all_module_hyperparams} shows their specific training hyperparameters.

\subsubsection{Rigid Object Segmentation Module} This module is dedicated to segmenting individual rigidly moving objects. We adapt the SOLOv2~\cite{wang2020solov2} architecture, modifying its input to utilize bidirectional optical flow. Given that the input is solely optical flow, which has a relatively low information density compared to RGB images, we employ a lightweight ResNet-18~\cite{ResNet} as the backbone to achieve strong performance efficiently.
\begin{itemize}
    \item \textbf{Input}: The input to this module is bidirectional optical flow. For an input video clip of length $t=7$ frames, we extract $t-1=6$ forward and $t-1=6$ backward optical flow fields. Each optical flow field has 2 channels (x and y components). Thus, the total number of input channels is $(t-1) \times 2 \times 2 = 24$. The input spatial resolution is $512 \times 512$ pixels.
    \item \textbf{Output}: The module outputs instance segmentation masks for rigidly moving objects within the central frame of the input clip.
\end{itemize}

\subsubsection{Motion Foreground Segmentation Module} This module is responsible for segmenting the entire motion foreground, which may include multiple rigid or non-rigid objects. It employs a U-Net~\cite{ronneberger2015unet} architecture for this binary segmentation task, also using a ResNet-18~\cite{ResNet} backbone, as the task of segmenting the entire motion foreground from optical flow is relatively straightforward.
\begin{itemize}
    \item \textbf{Input}: Similar to the rigid object segmentation module, the input is bidirectional optical flow. For a $t=7$ frame clip, this results in $(t-1) \times 2 \times 2 = 24$ input channels. The input spatial resolution is $512 \times 512$ pixels.
    \item \textbf{Output}: The module outputs a binary segmentation mask delineating the motion foreground for the central frame of the input clip.
\end{itemize}

\subsection{Perceptual Grouping Model (PGM)}
The Perceptual Grouping Model (PGM) is trained to internalize perceptual grouping principles from motion-derived pseudo-masks, enabling it to segment unseen objects in static images. This section details its training hyperparameters and inference strategy.

\subsubsection{PGM Training Details} Motion cues were extracted from a 10,000-hour video dataset using our MGMS pipeline. These cues, in the form of pseudo-masks, were subsequently filtered based on a mask quality score. This process yielded approximately 10,000,000 high-quality video frames, with an average of 2.1 pseudo-masks per frame. These frames and their corresponding motion-derived masks served as the training data for the PGM. The model was trained using a perceptual grouping contrastive learning strategy to learn the grouping logic inherent in these dynamic exemplars.

As described in the main paper, our PGM employs $K=4$ parallel heads to specialize in segmenting objects of varying scales. During training, each pseudo-mask is deterministically assigned to a specific head based on the size of its normalized bounding box. First, for each pseudo-mask, we compute its tightest bounding box and normalize its coordinates to the range of $[0, 1]$, resulting in coordinates $(x_{\text{min}}, y_{\text{min}}, x_{\text{max}}, y_{\text{max}})$. We then calculate a scale metric, $s$, defined as the geometric mean of the bounding box's side lengths:
\begin{equation}
    \label{eq:scale_metric}
    s = \sqrt{(x_{\text{max}} - x_{\text{min}}) \cdot (y_{\text{max}} - y_{\text{min}})}
\end{equation}
A pseudo-mask is then assigned to one of the four heads based on the value of $s$ according to the following criteria:
\begin{itemize}
    \item \textbf{Head 1}: Assigned masks with $s < \frac{1}{8}$ (small objects).
    \item \textbf{Head 2}: Assigned masks with $\frac{1}{8} \le s < \frac{1}{4}$ (medium-small objects).
    \item \textbf{Head 3}: Assigned masks with $\frac{1}{4} \le s < \frac{1}{2}$ (medium-large objects).
    \item \textbf{Head 4}: Assigned masks with $\frac{1}{2} \le s \le 1$ (large objects).
\end{itemize}
This scale-based routing ensures that the loss for any given mask is computed exclusively by the head designated for its scale, compelling each head to develop expertise in segmenting objects within its specific size range. The PGM was trained on a server with 8 NVIDIA A100 GPUs. Specific training hyperparameters are detailed in Table~\ref{tab:pgm_hyperparams}.

  \subsubsection{PGM Inference Strategy}\label{app:pgm_inference} The PGM is trained primarily on video frames, which often present inherent challenges such as lower resolutions (typically 360p to 720p) and significant motion blur. These characteristics can limit the effective resolution perceived by the model. Furthermore, the PGM's ViT-B/8 architecture, given a $512 \times 512$ pixel training input, generates segmentation maps at a coarse $64 \times 64$ resolution. To bridge this gap and enable high-resolution segmentation, especially for smaller objects and finer details, we implement a multi-scale tiling inference strategy. The procedure is as follows:

  \begin{enumerate}
      \item \textbf{Global Context Pass:} The entire input image is initially resized to $512 \times 512$ pixels and processed by the PGM. This step yields a coarse, global segmentation map.
      \item \textbf{Multi-Scale Sliding Window Pass:} The original, high-resolution image is subsequently processed using sliding tiles at two distinct scales:
      \begin{itemize}
          \item Scale 1: Tiles of size $s_1 \times s_1$, where $s_1 = \frac{1}{2} \min(\text{width}, \text{height})$.
          \item Scale 2: Tiles of size $s_2 \times s_2$, where $s_2 = \frac{1}{4} \min(\text{width}, \text{height})$.
      \end{itemize}
      These tiles are applied with a 50\% overlap to ensure comprehensive coverage and smooth transitions. The PGM performs an independent inference on each tile.
      \item \textbf{Aggregation and Refinement:} The segmentation masks from the global pass and all tiles are aggregated. Non-Maximum Suppression~\cite{neubeck2006efficient} is then applied to the combined predictions to resolve overlaps and redundant detections. As a final refinement step, we apply a fully-connected Conditional Random Field (CRF)~\cite{lafferty2001conditional} to the resulting masks. CRF leverages the color information from the original high-resolution image to sharpen mask boundaries, yielding a pixel-accurate final segmentation.
  \end{enumerate}

\subsection{Adaptation to Segment Anything}

To operationalize the learned perceptual grouping principles, we distill the knowledge from the PGM into two Segment Anything models: a whole-image segmentation model for automatic segmentation and a promptable segmentation model for interactive use.

\subsubsection{Whole-Image Segmentation Model} This model is designed for automatic, high-resolution segmentation. It employs a DINO pre-trained ResNet-50 backbone with a Mask2former decoder. To effectively manage the high density of perceptual grouping masks generated by the PGM, the decoder is configured with 1000 learnable object queries, though we sample a maximum of 400 masks per image during loss computation to ensure training efficiency. The model was trained for 8 epochs using the AdamW optimizer with a learning rate of $5 \times 10^{-5}$ and a batch size of 16, using random $1024 \times 1024$ crops for data augmentation.

\subsubsection{Promptable Segmentation model} This model is built for interactive use, generating hierarchical masks from user prompts. It is based on the Semantic-SAM framework and uses a self-supervised pretrained Swin-Tiny backbone. A key feature is its ability to output 6 masks for each prompt, capturing different levels of semantic granularity (e.g., a part, the whole object, and a group of objects). This design is inspired by the nested structure of real-world annotations. The model was trained for 4 epochs with a batch size of 8, using AdamW with a base learning rate of $1 \times 10^{-4}$ and a multi-step decay schedule.

\section{More Ablation Experiments}
\label{sec:more_ablation_experiments}

We provide additional analyses of the mask filtering strategy and video-based alternatives to further validate the design choices in MoSA.

\subsection{Mask Quality Filtering}
The mask quality score in our MGMS post-processing stage combines maskness and boundary sharpness. Table~\ref{tab:mask_quality_filtering} shows that the two criteria are complementary. Using maskness alone yields 24.1 average AR, while using boundary sharpness alone yields 22.8 average AR. Combining both scores improves the average AR to 28.8, confirming that reliable mask confidence and crisp boundaries are both important for constructing high-quality motion pseudo-labels.

\begin{table}[t]
    \centering
    \caption{Ablation of mask-quality filtering. We report AR on seven evaluation benchmarks and their average.}
    \label{tab:mask_quality_filtering}
    \small
    \resizebox{\textwidth}{!}{%
    \begin{tabular}{lcccccccc}
    \toprule
     & PtIn & LVIS & Entity & PACO & ADE & COCO & SA-1B & Average \\
    \midrule
    Only maskness & 28.6 & 24.3 & 24.5 & 13.9 & 21.2 & 27.6 & 28.9 & 24.1 \\
    Only sharpness & 26.3 & 22.7 & 23.5 & 12.8 & 20.0 & 27.1 & 27.2 & 22.8 \\
    Maskness + sharpness & \textbf{34.7} & \textbf{27.9} & \textbf{28.2} & \textbf{16.1} & \textbf{25.5} & \textbf{32.6} & \textbf{36.6} & \textbf{28.8} \\
    \bottomrule
    \end{tabular}%
    }
\end{table}

\subsection{Video-Based Alternatives}
We also compare MGMS with representative video-based unsupervised segmentation methods. OCLR~\cite{OCLR} and VideoCutLER~\cite{VideoCutLER} mainly target instance-level video segmentation, whereas MGMS is designed to extract multi-granularity motion masks for learning a transferable object prior. As shown in Table~\ref{tab:davis_video_segmentation}, MGMS achieves competitive zero-shot video segmentation performance on DAVIS2017, reaching 55.9 $\mathcal{J}\&\mathcal{F}$. This is close to VideoCutLER (57.3) and OCLR (55.1), even though MGMS and OCLR use only optical flow while VideoCutLER uses RGB appearance cues.

\begin{table}[t]
    \centering
    \caption{Zero-shot video segmentation on DAVIS2017.}
    \label{tab:davis_video_segmentation}
    \small
    \begin{tabular}{lccc}
    \toprule
     & $\mathcal{J}\&\mathcal{F}$ & $\mathcal{J}$ (Mean) & $\mathcal{F}$ (Mean) \\
    \midrule
    OCLR~\cite{OCLR} & 55.1 & 54.5 & 55.7 \\
    VideoCutLER~\cite{VideoCutLER} & 57.3 & 57.4 & 57.2 \\
    MGMS (ours) & 55.9 & 55.8 & 56.0 \\
    \bottomrule
    \end{tabular}
\end{table}

\begin{figure}[t]
    \centering
    \includegraphics[width=0.6\textwidth]{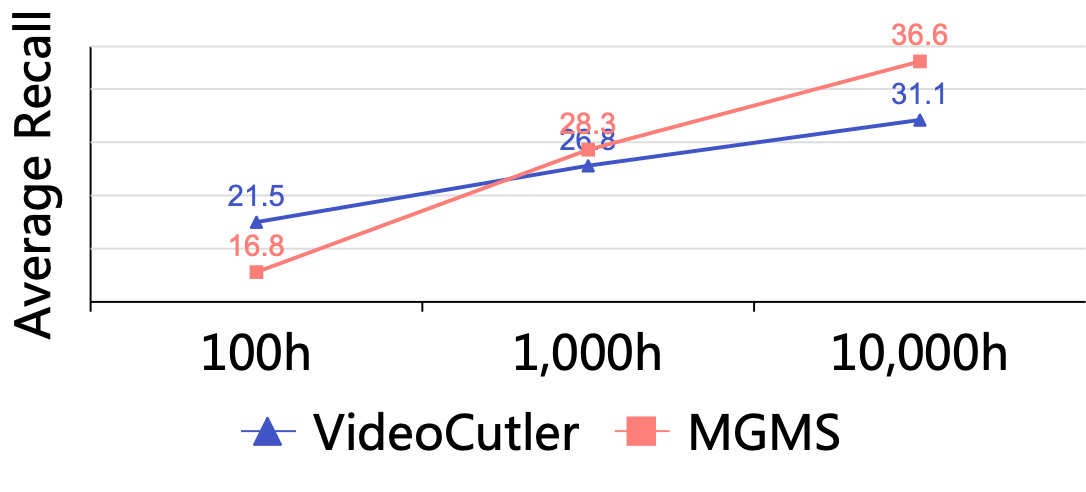}
    \caption{Scaling trend with video pre-training data. Compared with PGM trained using VideoCutLER pseudo-labels, PGM trained with our motion-driven pseudo-labels scales more robustly as video data increases.}
    \label{fig:video_scaling}
\end{figure}

Furthermore, we study how different pseudo-label sources scale with the amount of video pre-training data. When training PGM with VideoCutLER pseudo-labels, the reliance on DINO-based appearance features limits the scaling trend, likely because the pseudo-labels inherit static appearance biases. In contrast, our motion-driven pseudo-labels yield consistent improvements as the video pre-training data grows, as shown in Fig.~\ref{fig:video_scaling}.

\section{Additional Qualitative Results}
\label{app:qualitative_results}
\begin{figure*}[t]
    \centering
    \includegraphics[width=1\textwidth]{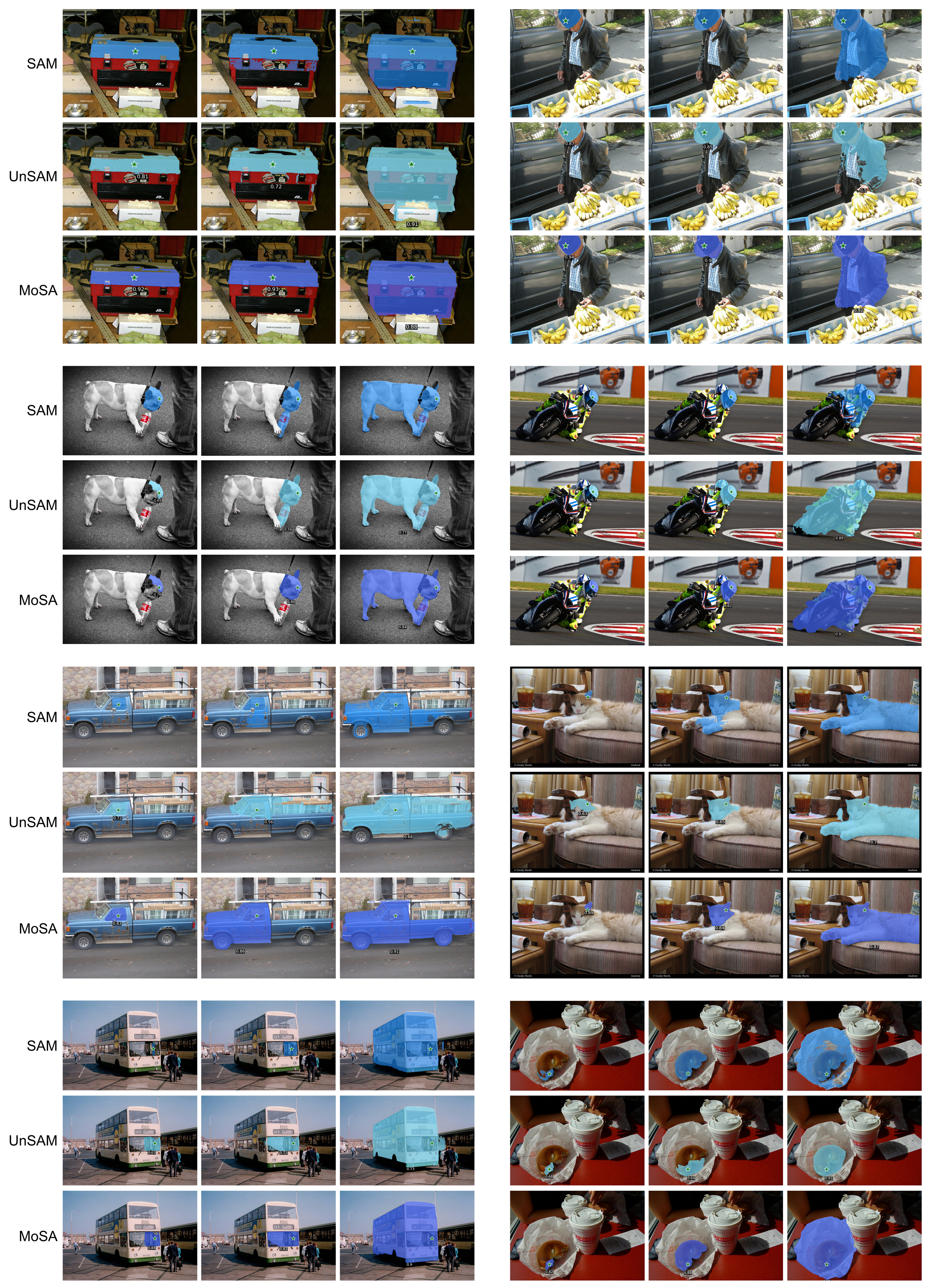}
    \caption{Comparative results for promptable image segmentation, evaluating our unsupervised MoSA against supervised SAM~\cite{SAM} and unsupervised UnSAM~\cite{UnSAM} across diverse scenario. The examples illustrate that when prompted by points (indicated as star marks), MoSA consistently generates multi-granular segmentation masks of higher visual quality. Note that UnSAM is not strictly unsupervised, as its refinement stage utilizes the supervised CascadePSP ~\cite{CascadePSP} refiner.}
    \label{promptable_segmentation}
\end{figure*}

We provide additional qualitative results demonstrating MoSA's segment anything capabilities in Fig.~\ref{whole_segmentation} and Fig.~\ref{promptable_segmentation} for whole-image segmentation and promptable image segmentation, respectively.

\section{Limitations}
\label{app:limitations}
Our current work focuses on extracting general segmentation knowledge from motion cues, with the goal of enabling zero-shot segmentation on static images without relying on explicit motion information during inference. Consequently, all evaluation experiments have been conducted on image datasets. Future research will extend MoSA to video data, exploring its applicability and adaptation to dynamic scenes and temporal segmentation tasks.

\begin{figure*}[t]
    \centering
    \includegraphics[width=1\textwidth]{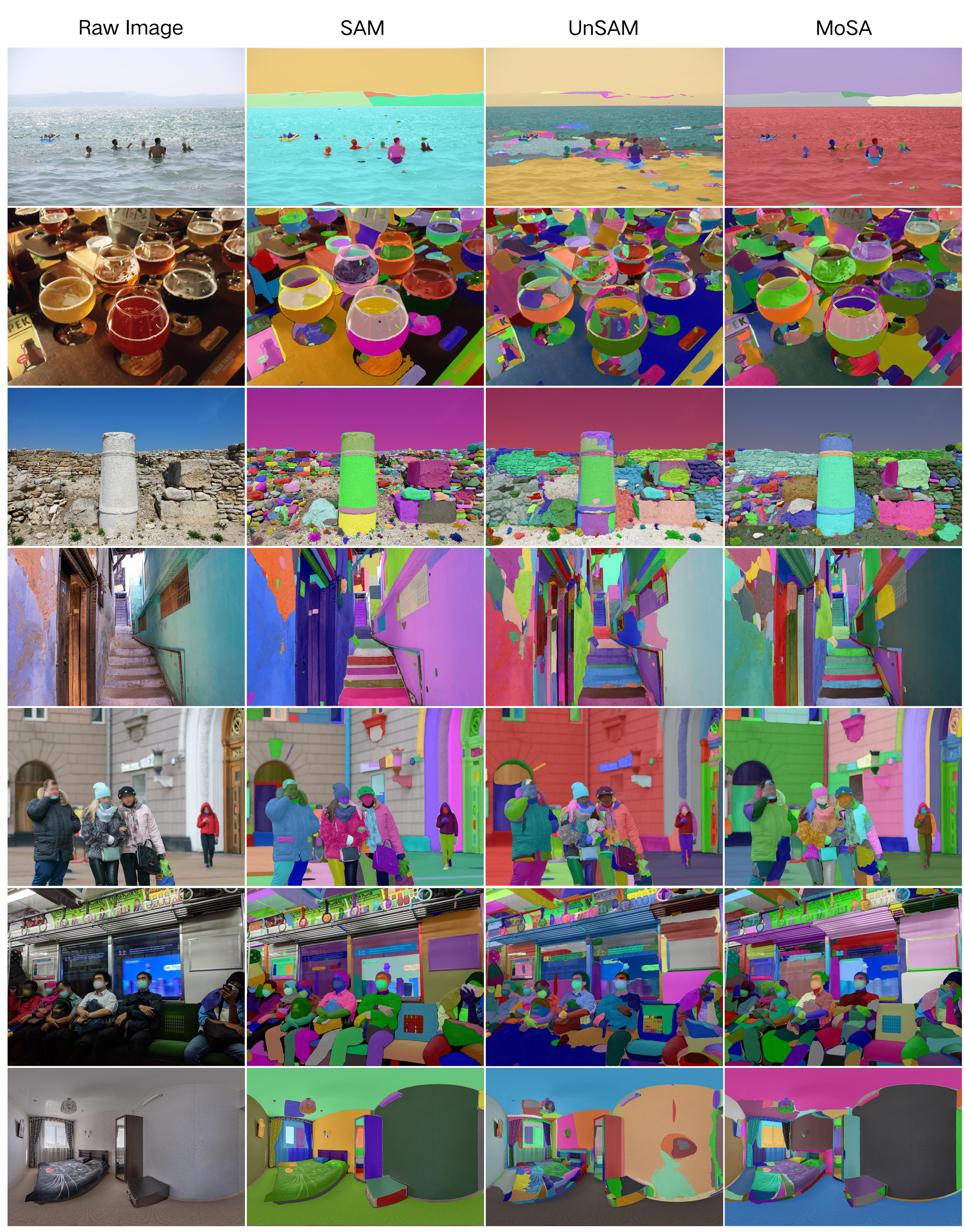}
    \caption{Additional qualitative results for whole-image segmentation, comparing our MoSA (rightmost column) with the supervised SAM~\cite{SAM} (second column) and the previous state-of-the-art unsupervised method UnSAM~\cite{UnSAM} (third column). Note that UnSAM is not strictly unsupervised, as its refinement stage utilizes the supervised CascadePSP ~\cite{CascadePSP} refiner.}
    \label{whole_segmentation}
\end{figure*}

One practical failure mode arises from degraded motion estimation in low-light environments, such as night scenes. Since our pseudo-label generation stage depends on optical flow, unreliable flow can fail to provide clear motion cues and may weaken the learned supervision for fine structures. For example, as shown in the penultimate row of Fig.~\ref{whole_segmentation}, MoSA can struggle to segment subtle body parts such as legs in dark regions, whereas a fully supervised model such as SAM can still succeed with dense manual supervision. Improving robustness to low-light videos and other flow-degraded conditions is an important direction for future work.


\end{document}